\documentclass[10pt,twocolumn,letterpaper]{article}

\usepackage{iccv}
\usepackage{times}
\usepackage{epsfig}
\usepackage{graphicx}
\usepackage{amsmath}
\usepackage{amssymb}

\usepackage{epstopdf}
\usepackage{multirow}
\usepackage{array}
\usepackage[linesnumbered,ruled]{algorithm2e}
\newcolumntype{L}[1]{>{\raggedright\let\newline\\\arraybackslash\hspace{0pt}}m{#1}}
\newcolumntype{C}[1]{>{\centering\let\newline\\\arraybackslash\hspace{0pt}}m{#1}}
\newcolumntype{R}[1]{>{\raggedleft\let\newline\\\arraybackslash\hspace{0pt}}m{#1}}
\usepackage[utf8]{inputenc}
\usepackage[english]{babel}
\usepackage{amsthm}
\usepackage[misc]{ifsym}

\usepackage[pagebackref=true,breaklinks=true,letterpaper=true,colorlinks,bookmarks=false]{hyperref}

\iccvfinalcopy 

\def\iccvPaperID{2132} 

\makeatletter
\newcommand{\printfnsymbol}[1]{%
  \textsuperscript{\@fnsymbol{#1}}%
}
\makeatother

\ificcvfinal\fi
\begin{document}

\title{Semantic Part Detection via Matching:\\Learning to Generalize to Novel Viewpoints from Limited Training Data}

\author{Yutong Bai\textsuperscript{1}\thanks{The first two authors contributed equally to this work.}, Qing Liu\textsuperscript{1}\printfnsymbol{1}, Lingxi Xie\textsuperscript{1,2,\Letter}, Weichao Qiu\textsuperscript{1}, Yan Zheng\textsuperscript{3}, Alan Yuille\textsuperscript{1}\\
\textsuperscript{1}Johns Hopkins University\quad\textsuperscript{2} Huawei Noah's Ark Lab\quad\textsuperscript{3}University of Texas at Austin\\
{\tt\small \{ytongbai, 198808xc, yan.zheng.mat, qiuwch, alan.l.yuille\}@gmail.com}\quad{\tt\small qingliu@jhu.edu}\\
}

\maketitle
\thispagestyle{empty}

\begin{abstract}

\textcolor{black}{Detecting semantic parts of an object is a challenging task, particularly because it is hard to annotate semantic parts and construct large datasets. 
In this paper, we present an approach which can learn from a small annotated dataset containing a limited range of viewpoints and generalize to detect semantic parts for a much larger range of viewpoints.
The approach is based on our matching algorithm, which is used for finding accurate spatial correspondence between two images and transplanting semantic parts annotated on one image to the other. 
Images in the training set are matched to synthetic images rendered from a 3D CAD model, following which a clustering algorithm is used to automatically annotate  semantic parts of the CAD model. 
During the testing period, this CAD model can synthesize annotated images under every viewpoint. 
These synthesized images are matched to images in the testing set to detect semantic parts in novel viewpoints. 
Our algorithm is simple, intuitive, and contains very few parameters. Experiments show our method outperforms standard deep learning approaches and, in particular, performs much better on novel viewpoints. For facilitating the future research, code is available: https://github.com/ytongbai/SemanticPartDetection}

\end{abstract}

\section{Introduction}

Detecting and parsing an object has been a long-lasting challenge in computer vision and has attracted a lot of research attention~\cite{everingham2010pascal,felzenszwalb2009object}. Recently, with the development of deep networks, this research area has been dominated by an approach which starts by extracting several regional proposals and then determines if each of them belongs to a specific set of object classes~\cite{girshick2014rich,ren2015faster,dai2016r,liu2016ssd,redmon2016you}. The success of these approaches~\cite{everingham2010pascal,lin2014microsoft} motivates researchers to address the more challenging task of understanding the objects at a finer level and, in particular, to parse it into semantic parts, which was defined to be those components of an object with semantic meaning and can be verbally described~\cite{zhang2018deepvoting}. A particular challenge lies in that annotating semantic parts is much more difficult and time-consuming than annotating objects, which makes it harder to directly apply deep networks to this task.

\newcommand{\figurewidth}{7cm}
\begin{figure}[!t]
\centering
\includegraphics[width=\figurewidth]{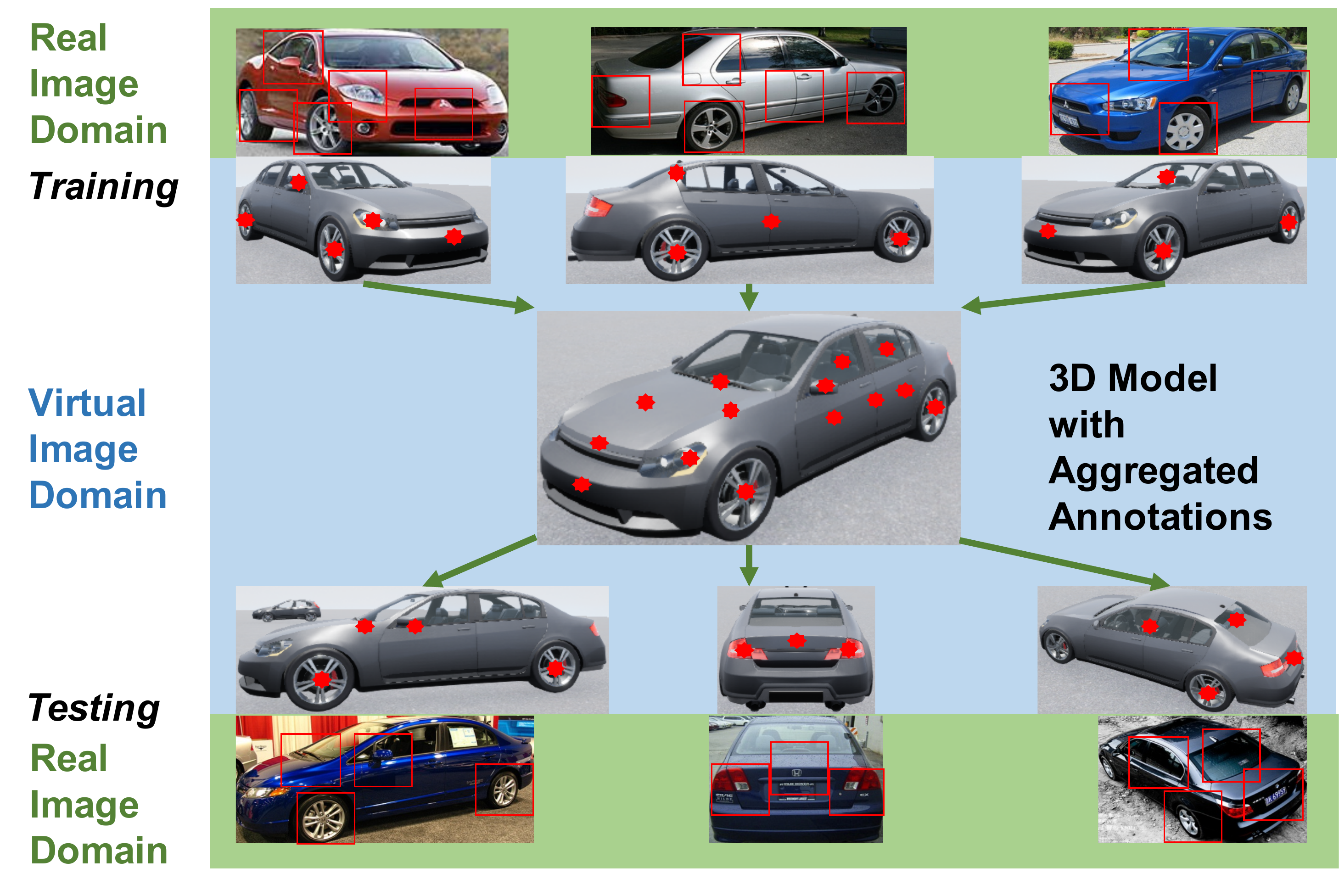}
\caption{The flowchart of our approach (best viewed in color). The key module is the matching algorithm which finds spatial correspondence between real and synthesized images in similar viewpoints. This enables us to match the training data to a 3D CAD model and thereby annotate it. The CAD model can then be used to detect the semantic parts on images in the testing set, by reusing the matching algorithm. }
\label{Fig:Framework}
\vspace{-0.4cm}
\end{figure}

In this paper, we address the problem of semantic part detection in the scenario that only a small amount of training data are available and the object is seen from a limited range of viewpoints. The overall framework is illustrated in Figure~\ref{Fig:Framework}. Our strategy is to design a matching algorithm which finds correspondences between images of the same object seen from roughly the same viewpoint. This can be used to match the training real images to the rendered images of a 3D CAD model, enabling us to annotate the semantic parts of the 3D model automatically. The same matching algorithm can then be used in the testing time, which transplants the annotated semantic parts on the CAD model to the testing images, even though their viewpoints may not have appeared in the training set.

In this pipeline, the key component is the matching algorithm. 
For simplicity, we only assume it to work on two images, one real and one synthesized, with similar viewpoints. The viewpoint of the real image is provided by ground-truth. Meanwhile, the synthesized image can be rendered using the viewpoint of the real image. On these two images, regional features are extracted using a pre-trained network, and matched using an optimization algorithm with geometric constraints considered. Our framework has potential to enable more accurate matching algorithms to be used in the future to improve the performance of object parsing.


The major contribution of this work is to provide a simple and intuitive algorithm for semantic part detection which works using little training data and can generalize to novel viewpoints. It is an illustration of how virtual data can be used to reduce the need for time-consuming semantic part annotation. Experiments are performed on the VehicleSemanticPart (VSP) dataset~\cite{wang2015unsupervised}, which is currently the largest corpus for semantic part detection. \textcolor{black}{Our approach achieves better performance than standard end-to-end methods such as Faster R-CNN~\cite{ren2015faster} and DeepVoting~\cite{zhang2018deepvoting} in {\em car}, {\em bicycle} and {\em motorbike}. The advantages become even bigger when the amount of training data is small.}

The remainder of this paper is organized as follows. Section~\ref{RelatedWork} briefly reviews the prior literature, and Section~\ref{Approach} presents our framework. After experiments are shown in Section~\ref{Experiments}, we conclude this work in Section~\ref{Conclusions}.

\section{Related Work}
\label{RelatedWork}

In the past years, deep learning~\cite{lecun2015deep} has advanced the research and applications of computer vision to a higher level. With the availability of large-scale image datasets~\cite{deng2009imagenet} as well as powerful computational devices, researchers designed very deep neural networks~\cite{krizhevsky2012imagenet,simonyan2014very,szegedy2015going} to accomplish complicated vision tasks. The fundamental idea of deep learning is to organize neurons (the basic units that perform specified mathematical functions) in a hierarchical manner, and tune the parameters by fitting a dataset. Based on some learning algorithms to alleviate numerical stability issues~\cite{nair2010rectified,srivastava2014dropout,ioffe2015batch}, researchers developed deep learning in two major directions, namely, increasing the depth of the network towards higher recognition accuracy~\cite{he2016deep,huang2017densely,hu2018squeeze}, and transferring the pre-trained models to various tasks, including feature extraction~\cite{donahue2014decaf,sharif2014cnn}, object detection~\cite{girshick2014rich,girshick2015fast,ren2015faster}, semantic segmentation~\cite{long2015fully,chen2017deeplab}, pose estimation~\cite{newell2016stacked}, boundary detection~\cite{xie2015holistically}, {\em etc}.

For object detection, the most popular pipeline, in the context of deep learning, involves first extracting a number of bounding-boxes named regional proposals~\cite{alexe2012measuring,uijlings2013selective,ren2015faster}, and then determining if each of them belongs to the target class~\cite{girshick2014rich,girshick2015fast,ren2015faster,dai2016r,liu2016ssd,redmon2016you}. To improve spatial accuracy, the techniques of bounding-box regression~\cite{jiang2018acquisition} and non-maximum suppression~\cite{hosang2017learning} were widely used for post-processing. Boosted by high-quality visual features and end-to-end optimization, this framework significantly outperforms the conventional deformable part-based model~\cite{felzenszwalb2009object} which were trained on top of handcrafted features~\cite{dalal2005histograms}. Despite the success of this framework, it still suffers from weak explainability, as both object proposal extraction and classification modules were black-boxes, and thus easily confused by occlusion~\cite{zhang2018deepvoting} and adversarial attacks~\cite{xie2017adversarial}. There were also research efforts of using mid-level or high-level contextual cues to detect objects~\cite{wang2015unsupervised} or semantic parts~\cite{zhang2018deepvoting}. These methods, while being limited on rigid objects such as vehicles, often benefit from better transferability and work reasonably well on partially occluded data~\cite{zhang2018deepvoting}.

Another way of visual recognition is to find correspondence between features or images so that annotations from one (training) image can be transplanted to another (testing) image~\cite{han2017scnet,kim2017fcss,novotny2017anchornet,thewlis2017unsupervised,ufer2017deep,liu2011nonparametric}. This topic was noticed in the early age of vision~\cite{okutomi1993multiple} and later built upon handcrafted features~\cite{matas2004robust,hosni2012fast,yang2014daisy}. There were efforts in introducing semantic information into matching~\cite{liu2010sift}, and also improving the robustness against noise~\cite{ma2015robust}. Recently, deep learning has brought a significant boost to these problems by improving both features~\cite{sharif2014cnn,zhou2014object} and matching algorithms~\cite{dosovitskiy2015flownet,ham2016proposal,zhou2016learning,ilg2017flownet,ufer2017deep}, while a critical part of these frameworks still lies in end-to-end optimizing deep networks.

Training a vision system requires a large amount of data. To alleviate this issue, researchers sought for help from the virtual world, mainly because annotating virtual data is often much easier and cheaper~\cite{qiu2016unrealcv}. Another solution is to perform unsupervised or weakly-supervised training with consistency that naturally exists~\cite{zhou2016learning,godard2017unsupervised,zhu2017unpaired}. This paper investigates both of these possibilities.

\section{Our Approach}
\label{Approach}

\subsection{Problem: Semantic Part Detection}
\label{Approach:Problem}

\renewcommand{\figurewidth}{8.5cm}
\begin{figure}[!t]
\centering
\includegraphics[width=\figurewidth]{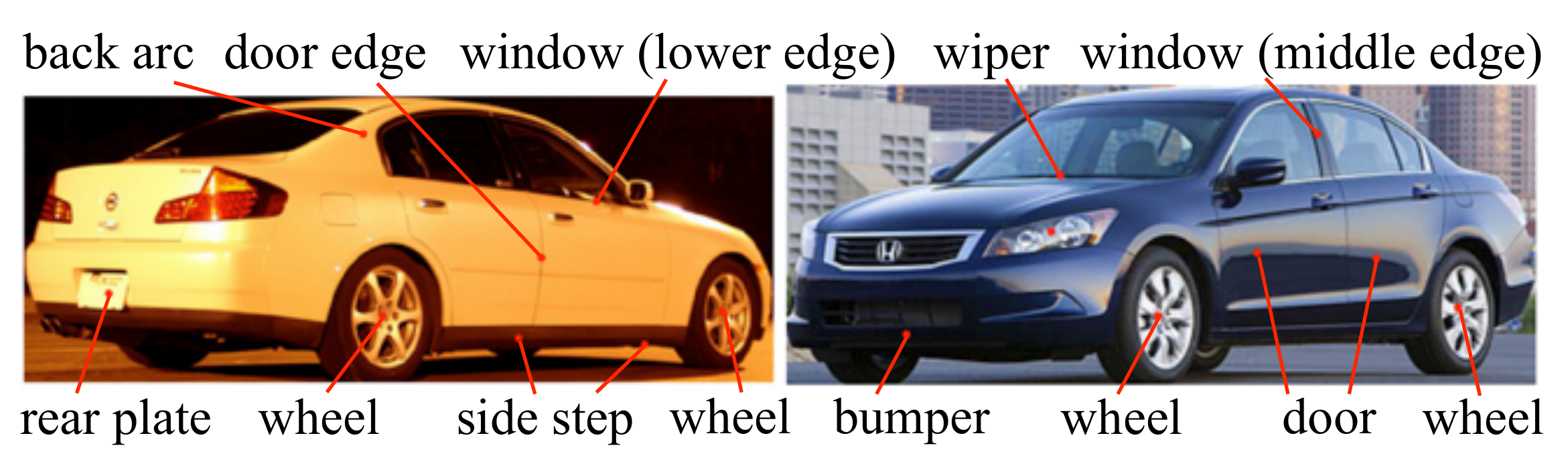}
\caption{Two examples of annotated semantic parts in the class {\em car}. For better visualization, we only show part of the annotations.}
\label{Fig:SemanticParts}
\vspace{-0.4cm}
\end{figure}

The goal of this work is to detect {\em semantic parts} on an image. We use $\mathcal{P}$ to denote the image-independent set of semantic parts, each element in which indicates a verbally describable component of an object~\cite{wang2015unsupervised}. For example, there are tens of semantic parts in the class {\em car}, including {\em wheel}, {\em headlight}, {\em license plate}, {\em etc}. Two {\em car} examples with semantic parts annotated are shown in Figure~\ref{Fig:SemanticParts}.

Let the training set $\mathcal{D}$ contain $N$ images, and each image, $\mathbf{I}_n$, has a spatial resolution of $W_n\times H_n$. A set $\mathcal{S}_n^\star$ of $M_n^\star$ semantic parts are annotated for each image, and each semantic part appears as a bounding box $\mathbf{b}_{n,m}^\star$ and a class label ${s_{n,m}^\star}\in{\left\{1,2,\ldots,\left|\mathcal{P}\right|\right\}}$, where $m$ is the index.

The goal of this work is to detect semantic parts in a testing image, in particular when the number of training images is very small, {\em e.g.}, $N$ is smaller than $50$.

\subsection{Framework: Detection by Matching}
\label{Approach:Framework}

We desire a function ${\mathcal{S}}={\mathbf{f}\!\left(\mathbf{I};\boldsymbol{\theta}\right)}$ which receives an image $\mathbf{I}$ and outputs the corresponding semantic part set $\mathcal{S}$. In the context of deep learning, researchers designed end-to-end models~\cite{ren2015faster} which start with an image, pass the signal throughout a series of layers, and output the prediction in an encoded form. With ground-truth annotation $\mathcal{S}_n^\star$, a loss function $\mathcal{L}_n$ is computed between ${\mathcal{S}_n}={\mathbf{f}\!\left(\mathbf{I}_n;\boldsymbol{\theta}\right)}$ and $\mathcal{S}_n^\star$, and the gradient of $\mathcal{L}_n$ with respect to $\boldsymbol{\theta}$ is computed in order to update $\boldsymbol{\theta}$ accordingly. DeepVoting~\cite{zhang2018deepvoting} went a step further by explicitly formulating spatial relationship at high-level inference layers so that the model can deal with partial occlusion. Despite their success, these approaches often require a large number of annotations to avoid over-fitting (see Section~\ref{Experiments:Results:TrainingData}), yet their ability to generalize to unseen viewpoints is relatively weak (see Section~\ref{Experiments:Diagnosis:UnseenData}). 

This paper works from another perspective which, instead of directly optimizing $\mathbf{f}\!\left(\cdot\right)$, adopts an indirect approach to find semantic correspondence between two images with similar viewpoints. This is to say, if a training image $\mathbf{I}_n$ is annotated with a set of semantic parts, $\mathcal{S}_n^\star$, and we know that $\mathbf{I}_n$ is densely matched to a testing image $\mathbf{I}_\circ$, then we can transplant $\mathcal{S}_n^\star$ to $\mathcal{S}_\circ$ by projecting each element of $\mathcal{S}_n^\star$ into the corresponding element of $\mathcal{S}_\circ$ by applying a spatially-constrained mapping function.

This method has two key factors. First, it works in the scenario of few ({\em e.g.}, tens of) training images. Second, it assumes that for every testing image, there exists a training image which has a very similar viewpoint (because the accuracy of semantic matching is not guaranteed in major viewpoint change). To satisfy these two conditions that seem to contradict, we make the {\bf main contribution} of this work, that introduces auxiliary cues from a 3D model $\mathbb{A}$. $\mathbb{A}$ is a virtual model created by computer graphics. Provided a viewpoint $\mathbf{p}$, a rendering function can generate an image $\mathbf{I}'$, possibly with semantic parts if they are present in $\mathbb{A}$.

Now we describe the overall flowchart as follows. In the training stage, we estimate the parameters of $\mathbb{A}$ ({\em e.g.}, 3D vertices, semantic parts, {\em etc.}) from the training set, for which we render $\mathbb{A}$ in the viewpoint of each training image and perform semantic matching. In the testing stage, we render $\mathbb{A}$ using the predicted viewpoint of the testing image, and make use of semantic matching to transplant the semantic parts to the testing image. Viewpoint prediction will be described in Section~\ref{Approach:ViewpointPrediction} in details.

In what follows, we formulate our approach mathematically and solve it using an efficient optimization algorithm.

\subsection{Semantic Matching and Viewpoint Consistency}
\label{Approach:Formulation}

This subsection describes three key modules (a rendering function, a semantic matching function, and a coordinate transformation function) and two loss terms (for geometric consistency and semantic consistency, respectively) that compose of the overall objective function.

We start with defining two key functions. First, a {\bf rendering function}, $\mathbf{R}\!\left(\cdot\right)$, takes the 3D model $\mathbb{A}$, refers to the target viewpoint $\mathbf{p}_n$, and outputs a rendered image:
\begin{equation}
\label{Eqn:ImageGeneration}
{\mathbf{I}_n'}={\mathbf{R}\!\left(\mathbb{A}\mid\mathbf{p}_n\right)}.
\end{equation}
Throughout the remaining part, a prime in superscript indicates elements in a generated image.

Next, we consider a {\bf semantic matching function} between a rendered image $\mathbf{I}_n'$ and a real image $\mathbf{I}_n$ of the same viewpoint\footnote{In the testing process, $\mathbf{I}_n$ and $\mathbf{I}_n'$ are replaced by $\mathbf{I}_\circ$ and $\mathbf{I}_\circ'$ (the 3D model rendered in the estimated viewpoint of $\mathbf{I}_\circ$), respectively.}. We assume that both $\mathbf{I}_n$ and $\mathbf{I}_n'$ are represented by a set of regional features, each of which describes the appearance of a patch. In the context of deep learning, this is achieved by extracting mid-level neural responses from a pre-trained deep network~\cite{wang2015unsupervised,zhou2014object}. Although it is possible to fine-tune the network with an alternative head for object detection~\cite{zhang2018deepvoting}, we do not take this option for simplicity. Let an image $\mathbf{I}_n$ have a set, $\mathcal{V}_n$, consisting of $L_n$ regional features, the $l$-th of which is denoted by $\mathbf{v}_{n,l}$. We assume that all these feature vectors have a fixed length, {\em e.g.}, all of them are $512$-dimensional vectors corresponding to specified positions at the {\em pool-4} layer of VGGNet~\cite{simonyan2014very,wang2015unsupervised}. Each $\mathbf{v}_{n,l}$ is also associated with a 2D coordinate $\mathbf{u}_{n,l}$. Based on these features, we build a matching function, $\mathbf{M}\!\left(\cdot\right)$, which takes the form of:
\begin{equation}
\label{Eqn:Matching}
{\mathcal{M}_n}={\mathbf{M}\!\left(\mathbf{I}_n,\mathbf{I}_n'\right)}={\left\{\left(l_{n,w},l_{n,w}'\right)\right\}_{w=1}^W},
\end{equation}
which indicates that the $l_{n,w}$-th feature in $\mathcal{V}_n$ and the $l_{n,w}'$-th feature in $\mathcal{V}_n'$ are matched. Based on the assumption that $\mathbf{I}_n$ and $\mathbf{I}_n'$ have similar viewpoints\footnote{In the testing process, we estimate the viewpoint of $\mathbf{I}_\circ$ and render a new image $\mathbf{I}_\circ'$ correspondingly. The estimator may bring in inaccuracy so that the viewpoints of these two images can be slightly different ({\em e.g.}, \cite{szeto2017click} reported a medium viewpoint prediction error of less than $10^\circ$ on the {\em car} subclass), but most often, this does not harm the matching algorithm.}, we make use of both unary and binary relationship terms to evaluate the quality of $\mathcal{M}_n$ in terms of appearance and spatial consistency, and thus define a {\bf semantic matching loss} of $\mathcal{M}_n$, $\alpha\!\left(\mathcal{M}_n\right)$:
\begin{eqnarray}
\nonumber
{\alpha\!\left(\mathcal{M}_n\right)}={\lambda_1\cdot{\sum_{w=1}^W}\left|\mathbf{v}_{n,l_{n,w}}-\mathbf{v}_{n,l_{n,w}'}'\right|^2+}\quad\quad\quad\quad\\
\label{Eqn:MatchingScore}
{\lambda_2\cdot{\sum_{1\leqslant w_1<w_2\leqslant W}}\left|\Delta\mathbf{u}_{l_{n,w_1},l_{n,w_2}}-\Delta\mathbf{u}_{n,l_{n,w_1}',l_{n,w_2}'}'\right|^2},
\end{eqnarray}
where $\Delta$ denotes the oriented distance between two feature coordinates, {\em i.e.}, ${\Delta\mathbf{u}_{l_{n,w_1},l_{n,w_2}}}={\mathbf{u}_{l_{n,w_1}}-\mathbf{u}_{l_{n,w_2}}}$ and ${\Delta\mathbf{u}_{n,l_{n,w_1},l_{n,w_2}}'}={\mathbf{u}_{n,l_{n,w_1}}'-\mathbf{u}_{n,l_{n,w_2}}'}$. Thus, the first term on the right-hand side measures the similarity in appearance of the matched patches, and the second term measures the spatial consistency among all matched patch pairs.

Based on $\mathcal{M}_n$, we can compute a {\bf coordinate transformation function}, $\mathbf{T}\!\left(\cdot\right)$, which maps the bounding box of each semantic part of $\mathbf{I}_n$ to the corresponding region on $\mathbf{I}_n'$:
\begin{equation}
\label{Eqn:Transformation}
{\mathbf{b}_{n,m}'}={\mathbf{T}\!\left(\mathbf{b}_{n,m}\mid\mathcal{M}_n\right)},\quad{s_{n,m}'}={s_{n,m}}.
\end{equation}
The annotations collected from all training images form a set, $\mathcal{B}$. Recall that the final goal is to infer the semantic parts of $\mathbb{A}$, which form a subset of vertices denoted by $\mathcal{C}$. To this end, we define the second loss term, a {\bf semantic consistency loss}, denoted by $\beta\!\left(\mathcal{B}\mid\mathcal{C}\right)$, which measures the inconsistent pairs of semantic parts on $\mathcal{B}$ and $\mathcal{C}$:
\begin{equation}
\label{Eqn:ConsistencyScore}
{\beta\!\left(\mathcal{B},\mathcal{C}\right)}={\lambda_3\cdot{\sum_{n=1}^N}{\sum_{m=1}^{M_n^\star}}\min_{s_{n,m}=c_{m^\ast}}\mathrm{dist}\!\left(\mathbf{b}_{n,m}',\mathbf{c}_{n,m^\ast}\right)+\lambda_4\cdot\left|\mathcal{C}\right|},
\end{equation}
where $\mathcal{C}$ has $M^\ast$ elements, each of which has a semantic part ID, $c_{m^\ast}$. $\mathbf{c}_{n,m^\ast}$ is the projected coordinate of the $m^\ast$-th semantic part in the viewpoint of $\mathbf{p}_n$. The distance between $\mathbf{b}_{n,m}'$ and $\mathbf{c}_{n,m^\ast}$, $\mathrm{dist}\!\left(\cdot,\cdot\right)$, is measured by the Euclidean distance between the centers. The total distance of matching together with a regularization term $\left|\mathcal{C}\right|$ contributes to the penalty.

The final objective is to minimize the overall loss function which sums up Eqns~\eqref{Eqn:MatchingScore} and~\eqref{Eqn:ConsistencyScore} together:
\begin{equation}
\label{Eqn:OverallScore}
{\mathcal{L}\!\left(\mathbb{A},\mathcal{D}\right)}={{\sum_{n=1}^N}\alpha\!\left(\mathcal{M}_n\right)+\beta\!\left(\mathcal{B},\mathcal{C}\right)}.
\end{equation}


\subsection{Training and Testing}
\label{Approach:Training}

Both training and testing involve optimizing Eqn~\eqref{Eqn:OverallScore}.

Training determines the semantic parts on $\mathbb{A}$ based on those annotated in $\mathcal{D}$, while testing applies the transplants the learned semantic parts from $\mathbb{A}$ to an unseen image. For simplicity, we assume that all the vertices of $\mathbb{A}$ are fixed, which means that a pre-defined 3D model is given. With more powerful 3D matching techniques in the future, this assumption can be relaxed, allowing the 3D model itself to be inferred by the dataset. Based on this, the overall optimization process is partitioned into four steps. During optimization, we replace parameters $\lambda_1$ through $\lambda_4$ with empirically set thresholds, as described below. Our algorithm is not sensitive to these parameters.

The {\bf first} step involves Eqn~\eqref{Eqn:ImageGeneration} which renders all virtual images $\mathbf{I}_n'$ according to the ground-truth viewpoint $\mathbf{p}_n$ (of $\mathbf{I}_n$). We purchased a high-resolution model from the Unreal Engine Marketplace, and use UnrealCV~\cite{qiu2016unrealcv}, a public library based on Unreal Engine, to synthesize all images. The rendering function $\mathbf{R}\!\left(\cdot\right)$ is implemented with standard rasterization in a game engine. We place the 3D model in regular background with {\em road} and {\em sky}, and use two directional light sources to reduce shadow in the rendered images (this improves image matching performance). Some typical examples are displayed in Figure~\ref{Fig:Framework}.

In the {\bf second} step, we match each training image $\mathbf{I}_n$ to the corresponding synthesized image $\mathbf{I}_n'$. This is to compute a function $\mathcal{M}_n$ that minimizes the semantic matching loss $\alpha\!\left(\mathcal{M}_n\right)$, as defined in Eqns~\eqref{Eqn:Matching} and~\eqref{Eqn:MatchingScore}. We use a fixed and approximate algorithm for this purpose. Given an image, either real or synthesized, we extract features by rescaling each image so that the short axis of the object is $224$-pixel long~\cite{wang2015unsupervised}, followed by feeding it into a pre-trained $16$-layer VGGNet~\cite{simonyan2014very} and extracting all $512$-dimensional feature vectors at the {\em pool-4} layer. All feature vectors are $\ell_2$-normalized. There are $L_n\times L_n'$ feature pairs between $\mathbf{I}_n$ and $\mathbf{I}_n'$. For each of them, $\left(l,l'\right)$, we compute the $\ell_2$ distance between $\mathbf{v}_{n,l}$ and $\mathbf{v}_{n,l'}'$ and use a threshold $\xi$ to filter them. On the survived features, we further enumerate all quadruples $\left(l_1,l_2,l_1',l_2'\right)$ with $l_1$ matched to $l_1'$ and $l_2$ matched to $l_2'$, compute $\Delta\mathbf{u}_{l_1,l_2}$ and $\Delta\mathbf{u}_{l_1',l_2'}'$, and again filter them with a threshold $\zeta$. Finally, we apply the Bron-Kerbosch algorithm to find the max-clique on both images that are matched with each other. An example of the matching algorithm is shown in Figure~\ref{Fig:Matching}. In practice, $\xi$ and $\zeta$ are determined empirically, and the matching performance is not sensitive to these parameters.

\renewcommand{\figurewidth}{7cm}
\begin{figure}[!t]
\centering
\includegraphics[width=\figurewidth]{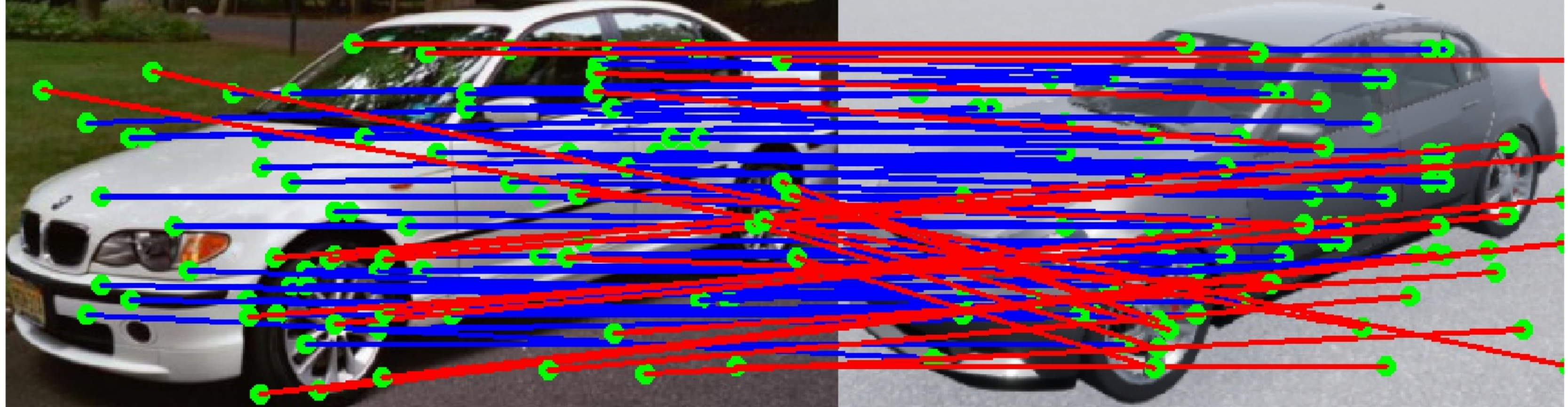}\\
\includegraphics[width=\figurewidth]{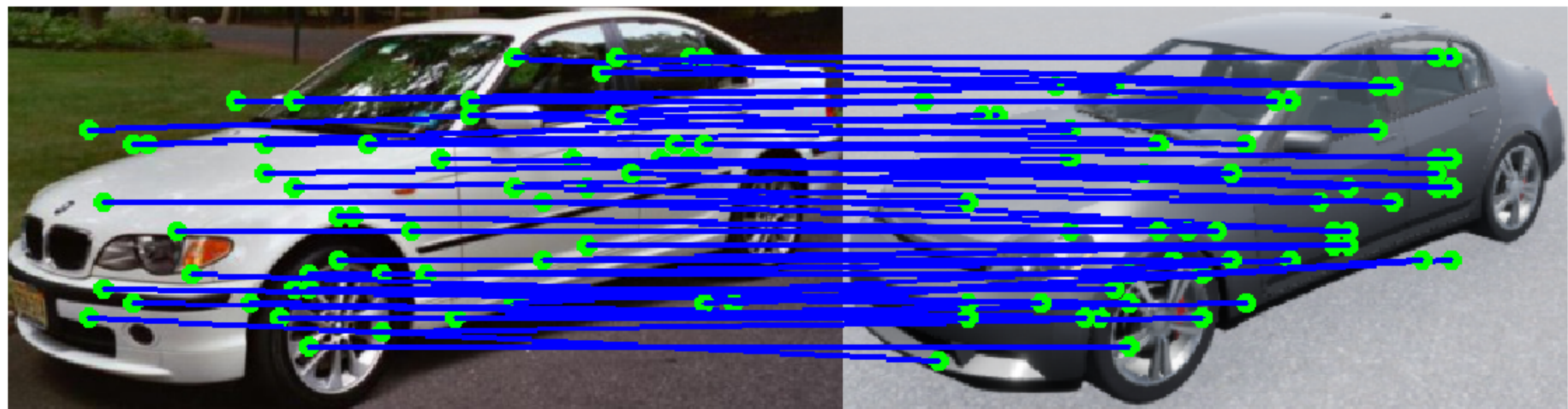}\\
\includegraphics[width=\figurewidth]{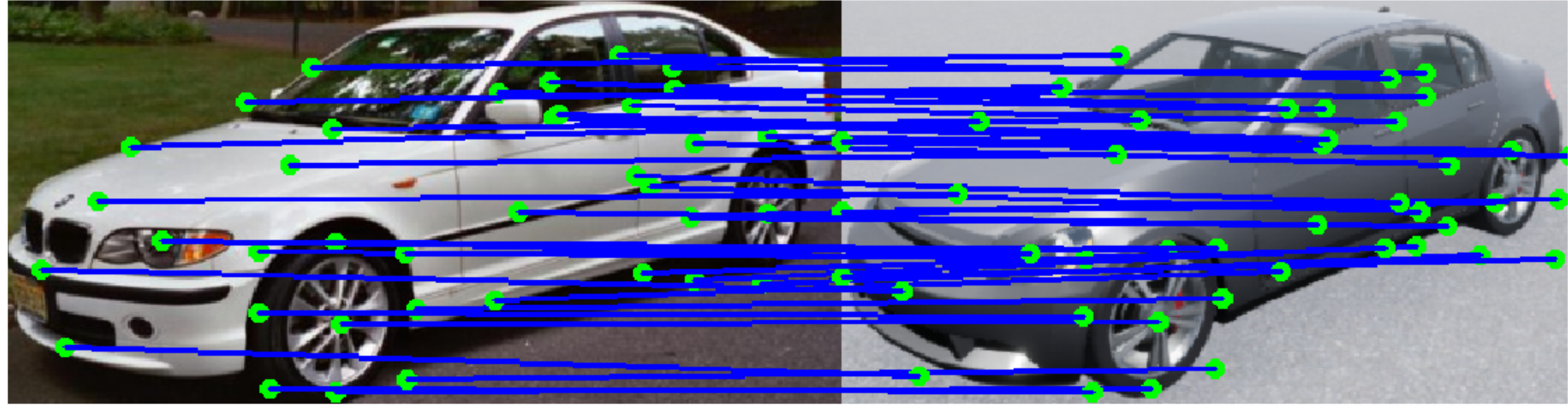}
\caption{\textcolor{black}{The matching process between real and synthesized images. The first row depicts the result using the matched regional features. The second represents the filtered matching pairs with geometric constraints. The third is the final matched key points after using {\em pool-3} features to refine.}
\vspace{-0.4cm}
}

\label{Fig:Matching}
\end{figure}

In the {\bf third} step, we compute the transformation function $\mathbf{T}\!\left(\cdot\right)$ defined in Eqn~\eqref{Eqn:Transformation}. This is to determine how each coordinate in one image is mapped to that in the other. We use the nearby matched features to estimate this function. For each semantic part, a weighted average of its neighboring features' relative translation is applied to the annotation, where the weights are proportional to the inverse of the 2D Euclidean distances between the semantic part and the features in the source image. The basis of our approach is the feature vectors extracted from a pre-trained deep network. However, these features, being computed at a mid-level layer, often suffer a lower resolution in the original image plane. For example, the {\em pool-4} features of VGGNet~\cite{simonyan2014very} used in this paper have a spatial stride of $16$, which leads to inaccuracy in feature coordinates and, consequently, transformed locations of semantic parts. To improve matching accuracy, we extract two levels of regional features. The idea is to first use higher-level ({\em e.g.}, {\em pool-4}) features for semantic matching, and then fine-tune the matching using lower-level ({\em e.g.}, {\em pool-3}) features which have a smaller spatial stride for better alignment.

The {\bf fourth} and final step varies from training and testing. In the {\em training} stage, we collect semantic parts from annotated real images, transform them to the virtual model, and determine the set $\mathcal{C}$ by minimizing Eqn~\eqref{Eqn:ConsistencyScore}. This is approximately optimized by a standard K-Means clustering on all transformed semantic parts, {\em i.e.}, the set of $\mathcal{B}$. After clustering is done, we enumerate the number of semantic parts, {\em i.e.}, $\left|\mathcal{C}\right|$, and choose the maximum clustering centers accordingly. This is to say, the final position of each semantic part is averaged over all transformed semantic parts that are clustered to the same center. \textcolor{black}{In practice, we take the center of the bounding box annotations and find the closest viewable vertex on the 3D model.} In the {\bf testing} stage, the reversed operation is performed, transforming the learned semantic parts back to each real image individually. This can introduce 3D prior to overcome the issue of occlusion (see experiments). 

\newcommand{\colwidthA}{1.0cm}
\begin{table*}
\centering{
\setlength{\tabcolsep}{0.08cm}
\small
\begin{tabular}{|l||R{\colwidthA}|R{\colwidthA}|R{\colwidthA}|R{\colwidthA}||R{\colwidthA}|R{\colwidthA}|R{\colwidthA}|R{\colwidthA}||R{\colwidthA}|R{\colwidthA}|R{\colwidthA}|R{\colwidthA}|}
\hline
\multirow{2}{*}{Approach} &
\multicolumn{4}{c||}{$16$ Training Samples} &
\multicolumn{4}{c||}{$32$ Training Samples} &
\multicolumn{4}{c|}{$64$ Training Samples} \\
\cline{2-13}
{}          &         {\bf L0} &         {\bf L1} &         {\bf L2} &         {\bf L3}
            &         {\bf L0} &         {\bf L1} &         {\bf L2} &         {\bf L3}
            &         {\bf L0} &         {\bf L1} &         {\bf L2} &         {\bf L3} \\
\hline
Faster R-CNN~\cite{ren2015faster}
            &          $31.64$ &          $15.57$ &          $10.51$ &          $8.35$
            &          $43.05$ &          $20.72$ &          $14.81$ &          $11.95$
            &          $\mathbf{55.43}$ &          $29.03$ &          $19.05$ &          $13.75$ \\
\hline
DeepVoting~\cite{zhang2018deepvoting}
            &          $33.28$ &          $18.09$ &          $13.92$ &          $10.26$
            &          $37.66$ &          $21.91$ &          $16.35$ &          $12.12$
            &          $49.23$ &          $30.01$ &          $21.86$ &          $15.52$ \\
\hline\hline
Ours        & $\mathbf{42.25}$ & $\mathbf{27.44}$ & $\mathbf{23.87}$ & $\mathbf{14.03}$
            & $\mathbf{44.06}$ & $\mathbf{28.29}$ & $\mathbf{23.76}$ & $\mathbf{14.58}$
            & $45.39$ & $\mathbf{33.93}$ & $\mathbf{25.24}$ & $\mathbf{16.75}$ \\
\hline
\end{tabular}}
\caption{\textcolor{black}{Semantic part detection accuracy (by mAP, $\%$) of different approaches under different number of training samples and occlusion situations. {\bf L0} through {\bf L3} indicate occlusion levels, with {\bf L0} being non-occlusion and {\bf L3} the heaviest occlusion.}} 
\label{Tab:Results}
\vspace{-0.4cm}
\end{table*}

\subsection{Similarity-based Viewpoint Prediction}
\label{Approach:ViewpointPrediction}

From diagnostic experiments, we can see that the accuracy of semantic part detection relies on that of viewpoint estimation. Except from using either ground-truth or an off-the-shelf method such as Click-Here-CNN~\cite{szeto2017click}, here we present an approach of using the quality of semantic matching itself to measure whether two images are similar enough, and thus provide an alternative to viewpoint estimation.

We start with the maximum clique computed on the matched features between two images. For the details of feature matching, graph construction and maximum clique computation, please refer to Section~3.4.

After the maximum clique algorithm, we obtain a shared sub-graph which is composed of the matched parts in two images. Let $N$ and $M$ be the number of vertices and edges in the graph defined by the maximum clique. Based on this graph, we design an energy function to describe the viewpoint similarity: 
\begin{equation}
\label{Eqn:EnergyFunc}
{E=\lambda\sum_{n=1}^{N}V_{n}+\frac{\mu}{N}\sum_{n=1}^{N}d_{n}+\frac{\gamma}{M}\sum_{m=1}^{M} \cos\left\langle\mathbf{v}_\mathbf{A}^m,\mathbf{v}_\mathbf{B}^m\right\rangle}.
\end{equation}
This is to say, the quality of matching is determined by the similarity of matched parts, their absolute position and their relative position. The {\bf first} term represents similarity of the parts: the similarity of the corresponding parts in two images, $V_{n}$, is measured by the inner-product of their features, after being normalized. Note that by summing up all feature pairs, we are actually taking the number of the matched parts into consideration, {\em i.e.}, the more matched feature pairs, the higher similarity. The {\bf second} term restricts the absolute position of these features: $d_n$ is the Euclidean distance between the 2D coordinates of two matched features (of two objects) in the cropped images. This is based on an intuitive assumption that, for example, if two cars have the same viewpoint, their corresponding key points should be located in approximately the same position in the image. The {\bf third} term confines the relative position of these parts by calculating the cosine distance $\cos\left\langle\mathbf{v}_{i},\mathbf{v}_{j}\right\rangle$ of the adjacent pairs of matched features. For a pair of features, $\mathbf{p}$ and $\mathbf{q}$, we calculate the cosine distance between $\mathbf{v}_\mathbf{A}$ and $\mathbf{v}_\mathbf{B}$, which point from $\mathbf{p}$ to $\mathbf{q}$ in $\mathbf{A}$ and $\mathbf{B}$. $\lambda$, $\mu$ and $\gamma$ are parameters to balance different terms.

Based on this function, the matching algorithm works as follows. Given a real image in the testing stage, we first enumerate $8$ synthetic images with the azimuth angles of $\{0,45,90,…315\}$. After finding the most similar case, we narrow down the searching range to a region of $90$ degrees, and enumerate at every $10$ degrees to find the best match. $10$ degrees is the amount that preserves most feature matches.

\subsection{Discussions}
\label{Approach:Discussions}

Compared to prior methods on object detection~\cite{ren2015faster} or parsing~\cite{zhang2018deepvoting}, our approach enjoys a higher explainability as shown in experiments (see Figure~\ref{Fig:interpret}). Here, we inherit the argument that low-level or mid-level features can be learned by deep networks as they often lead to better local~\cite{yi2016lift} or regional~\cite{sharif2014cnn} descriptions, but the high-level inference stage should be semantically meaningful so that we can manipulate either expertise knowledge or training data for improving recognition performance or transferring the pipeline to other tasks. Moreover, our approach requires much fewer training data to be optimized, and applies well in novel viewpoints which are not seen in training data.

The training process of our approach can be largely simplified if we fix $\mathbb{S}$, {\em e.g.}, manually labeling semantic parts on each 3D model $\mathbb{I}$. However, the amount of human labor required increases as the complexity of annotation as well as the number of 3D models. Our approach serves as a good balance -- the annotation on each 2D image can be transferred to different 3D models. In addition, 2D images are often annotated by different users, which provide complementary information by crowd-sourcing~\cite{deng2009imagenet}. Therefore, learning 3D models from the ensemble of 2D annotations is a safer option.

\section{Experiments}
\label{Experiments}

\textcolor{black}{
In experiments, we first compare our performance on detecting semantic parts of {\em sedan} with Faster-RCNN and DeepVoting. In this comparison, we studied the impact of the amount of data and occlusion level. Then, we explore the potential of our approach in various challenges, including its sensitivity to viewpoint prediction error, ability of working on unseen viewpoints, and ability of being applied to other prototypes.
Finally, we apply our model to two more vehicle types, namely {\em bicycle} and {\em motorbike}, which have fewer available training data.}

\subsection{Settings and Baselines}
\label{Experiments:SettingsBaselines}

We perform experiments on the VehicleSemanticPart (VSP) dataset~\cite{wang2015unsupervised}. 
We start with {\em sedan} the most popular prototype of {\em car}, then later shows only using a {\em sedan} CAD model, our approach can generalize to other prototypes ({\em e.g.}, {\em minivan}) without the need of a CAD model for each prototype.
There are $395$ training and $409$ testing images, all of which come from the Pascal3D+ dataset~\cite{xiang2014beyond}, and the authors of~\cite{wang2015unsupervised} manually labeled $39$ semantic parts, covering a large fraction of the surface of each {\em car} (examples in Figure~\ref{Fig:SemanticParts}). 
There are $9$ semantic parts related to {\em wheel}, $1$ at the center and other $8$ around the rim. We only consider the center one as the others are less consistent in the annotation.

We use the ground-truth azimuth angle to categorize training images into $8$ bins, centered at $0^\circ,45^\circ,\ldots,315^\circ$, respectively. We randomly sample ${N'}\in{\left\{2,4,8\right\}}$ images in each bin, leading to three training sets with $16$, $32$ and $64$ images, which are much smaller than the standard training set ($395$ images). Following the same setting of~\cite{zhang2018deepvoting}, we evaluate our approach on both clean object and occluded objects, where different levels of occlusion are tested.

The competitors of our approach include DeepVoting~\cite{zhang2018deepvoting}, a recent approach towards explainable semantic part detection, as well as Faster R-CNN~\cite{ren2015faster} (also used in~\cite{zhang2018deepvoting}), an end-to-end object detection algorithm. Other approaches ({\em e.g.}, \cite{wang2015unsupervised} and~\cite{wang2017detecting}) are not listed as they have been verified weaker than DeepVoting.

\subsection{Quantitative Results}
\label{Experiments:Results}

\textcolor{black}{We first investigate the scenario that the ground-truth viewpoint (quantized into the $8$ bins) is provided. Though obtaining extra information, we point out that annotating the viewpoint of an object often requires less than $3$ seconds -- in comparison, labeling all semantic parts costs more than one minute.} In later experiments, we also provide parsing results with predicted viewpoints and study the sensitivity of our approach to viewpoint accuracy.

Results are summarized in Table~\ref{Tab:Results}. One can observe that our approach outperforms both DeepVoting and Faster R-CNN, especially in the scenarios of (i) fewer training data and (ii) heavier occlusion.

\textcolor{black}{Since our approach is viewpoint-aware, we also equip baseline methods with ground-truth viewpoint information. More specifically, we train $8$ individual models (for both Faster R-CNN and DeepVoting), each of which takes charge of objects within one bin. On the clean testing images ({occlusion level L0}), the accuracy of Faster R-CNN is almost unchanged, and that of DeepVoting is improved by $1\%$--$6\%$, but still much lower than our results.}

\subsubsection{Impact of the Amount of Training Data}
\label{Experiments:Results:TrainingData}

One major advantage of our method is the ability to learn from just a few training samples by preserving the 3D geometry consistency for images from different viewpoints. As shown in Table~\ref{Tab:Results}, when using only $16$ training images, which means no more than $6$ training samples for most semantic parts, our method still gives reasonable predictions and outperforms other baseline method by a large margin. By increasing the training sample number, our method also benefits from learning more accurate annotations on the 3D model, resulting to higher mAP in the 2D detection task. By contrast, both Faster R-CNN and DeepVoting fail easily given small number of training.

\subsubsection{Ability of Dealing with Occlusion}
\label{Experiments:Results:Occlusion}
To evaluate the robustness of our method to occlusion, we apply the models learned from the occlusion-free dataset to images with different levels of occlusion. Different from DeepVoting~\cite{zhang2018deepvoting} which learns the spatial relationship between semantic parts and their characteristic deep features in occlusion-free images, our method directly models the spatial relationship of parts by projecting them to the 3D space, and then matches them back to the occluded objects. On light occlusions, our method consistently beats the baselines. In the scenarios of heavier occlusion, due to the deficiency of accurately matched features, the performance of our method deteriorates. As expected, Faster R-CNN lacks the ability of dealing with occlusion and its performance drops quickly, while DeepVoting is less affected.

It is interesting to see that the robustness to fewer training data and occlusion is negatively related to the number of extra parameters. For example, DeepVoting has less than $10\%$ parameters compared to Faster R-CNN, and our approach, being a stage-wise one, only requires some hyper-parameters to be set. This largely alleviates the risk of over-fitting (to small datasets) and the difficulty of adapting a model trained on non-occluded data to occluded data.



\subsubsection{Robustness on Unseen Viewpoints}

\label{Experiments:Diagnosis:UnseenData}

\begin{table}
\centering{
\small
\setlength{\tabcolsep}{0.08cm}
\begin{tabular}{|l||C{\colwidthA}|C{\colwidthA}|C{\colwidthA}|}
\hline
\multirow{2}{*}{Approach} & \multicolumn{3}{c|}{\# Training Samples} \\
\cline{2-4}
{}          &   $16$  & $32$  & $64$  \\
\hline
Faster R-CNN  &     $16.02$      &     $21.80$      &      $19.91$     \\
\hline
DeepVoting  &     $8.59$      &     $27.71$      &     $33.82$       \\
\hline \hline
Ours        & $\mathbf{45.32}$ & $\mathbf{47.03}$ & $\mathbf{45.88}$  \\
\hline
\end{tabular}}
\caption{\textcolor{black}{Results (by mAP, $\%$) of applying models trained with images under the viewpoint of an elevation angle $0^\circ$ to unseen viewpoints (an elevation angle $20^\circ$). Our model can generalize better to unseen viewpoints and the performance is almost constant regardless of the number of training data.}}
\label{Tab:Novel_VP}
\vspace{-0.4cm}
\end{table}

\renewcommand{\figurewidth}{7cm} 
\begin{figure}
\centering
\includegraphics[width=\figurewidth]{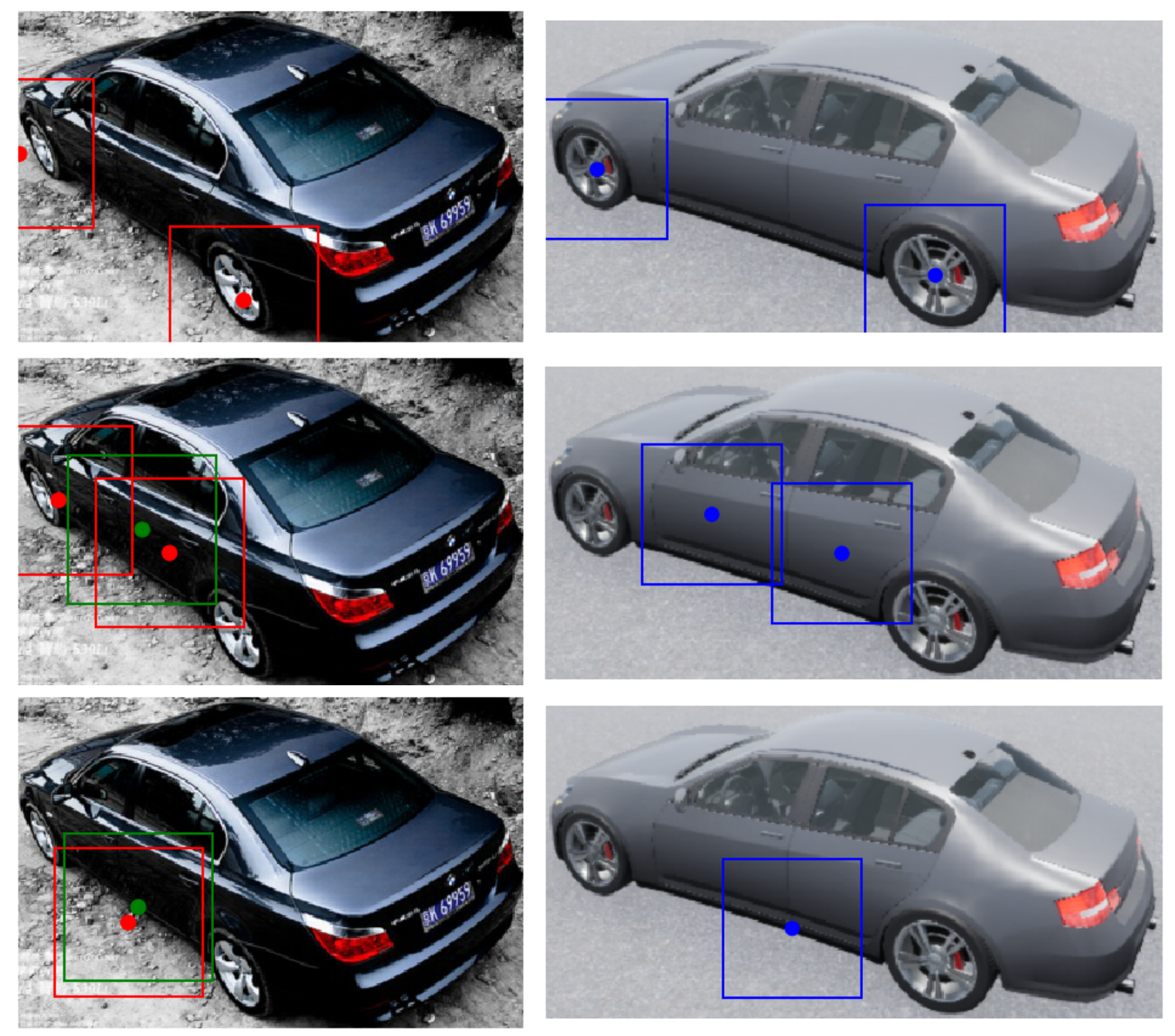}
\caption{\textcolor{black}{Our approach also works on unseen viewpoints (training on an elevation angle of $0^\circ$ and testing on $20^\circ$). It is worth noting that the {\em wheel} has been completely self-occluded in the elevation angle of $0^\circ$, but our model can still detect it. Left: a testing image with transferred annotations (red) and ground-truth (green). The ground-truth for {\em wheels} is missing, which is an annotation error. Right: the synthetic image in the same viewpoint with learned semantic parts. 
} }

\label{Fig:Novel_VP}
\end{figure}

To show that our approach has the ability of working on unseen viewpoints, we train the models using {\em sedan} images with various azimuth angle and $0^\circ$ elevation angle, then test them on {\em sedan} with elevation angle equal or larger than $10^\circ$. The results are shown in Table~\ref{Tab:Novel_VP}. Our method maintains roughly the same mAP as tested on all viewpoints, while Faster R-CNN and DeepVoting deteriorate heavily. In Figure~\ref{Fig:Novel_VP}, we show predictions made by our method on one sample with unseen viewpoint (elevation angle equals $20^\circ$). The predicted locations are very close to the annotated ground truth and may help to fix the annotation error in the dataset.

\subsubsection{Transfer Across Different Prototypes}
\label{Experiments:Diagnosis:Prototypes}

\begin{table}
\centering{
\small
\setlength{\tabcolsep}{0.08cm}
\begin{tabular}{|l||R{\colwidthA}|R{\colwidthA}|R{\colwidthA}|R{\colwidthA}|}
\hline
Approach          &         {\em sedan} &         {\em SUV} &         {\em minivan} &         {\em hatchback}\\
\hline
Faster R-CNN  &          $43.05$ &          $41.16$ &          $32.18$ &          $30.04$ \\
\hline
DeepVoting  &          $37.66$ &          $37.18$ &          $30.67$ &          $31.62$ \\
\hline \hline
Ours        & $\mathbf{47.27}$ & $\mathbf{42.80}$ & $\mathbf{35.58}$ & $\mathbf{32.02}$ \\
\hline
\end{tabular}}
\caption{\textcolor{black}{Results of applying a model trained with {\em sedan} images to other prototypes ({\em SUV}, {\em minivan}, and {\em hatchback}). Our model can generalize better to unseen prototypes.}}
\label{Tab:Prototype}
\vspace{-0.4cm}
\end{table}

In order to evaluate how sensitive our method is to the {\em car} prototype used during training, we transfer the model trained with {\em sedan} images to other prototypes of {\em cars}, including {\em minivan}, {\em SUV} and {\em hatchback}.
Results are summarized in Table~\ref{Tab:Prototype}. 
As expected, our method generalizes well to prototypes with similar appearance (e.g., {\em SUV}). For {\em minivan} and {\em hatch-back}, due to the variation of the 3D structures and semantic part definitions, the performance drops more. Similar results are observed for Faster R-CNN and DeepVoting, and DeepVoting seems slightly more robust to the prototypes.

\subsubsection{Extending to Other Vehicle Types}
\label{Experiments:Results:Objects}

\textcolor{black}{For vehicle, most of computer vision research focused on {\em car}, which has a lot of annotated images. {\em Bicycle} ,and {\em motorbike}, while being equally important objects on the road, attracted much fewer attention, partially due to the lack of data. The ability of training with few samples enables our model to be easily extended to these unrepresented classes. The accuracies of our approach for {\em bicycle} and {\em motorbike} are $67.16\%$ and $42.81\%$, respectively. Due to the lack of training data for these categories, although we can manage to train an end-to-end model, the performance is inevitably worse than us. Faster R-CNN reports average accuracies of $44.02\%$ and $29.79\%$ for {\em bicycle} and {\em motorbike}, and the numbers for DeepVoting are $59.43\%$ and $31.88\%$, respectively. This shows the practical advantage compared to prior work~\cite{ozuysal2009pose,glasner2012aware,li2017deep} which only focused on {\em car} parsing.
}

\subsection{Qualitative Studies}
\label{Experiments:Qualitative}

\subsubsection{Viewpoint Consistency in Training}
\label{Experiments:Qualitative:ViewpointConsistency}

\renewcommand{\figurewidth}{8cm}
\begin{figure}
\centering
\includegraphics[width=8.5cm]{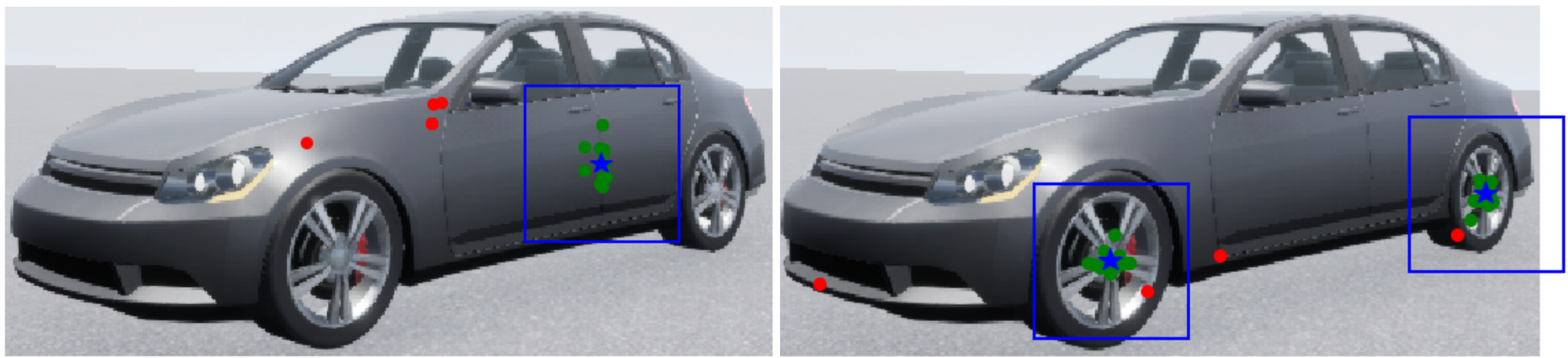}
\caption{Two examples of how viewpoint consistency improves the semantic part annotation on the 3D model. The red dots represent incorrectly transferred semantic part annotations that get eliminated during our aggregation process using 3D geometry constraints. The green dots are the reasonable annotations that are used to get the final annotation for the targeted semantic part, which is represented by the blue dots at the center of blue bounding-boxes.}
\label{Fig:vp_consistency}
\vspace{-0.4cm}
\end{figure}

In Figure~\ref{Fig:vp_consistency}, we show examples of how viewpoint consistency improves the stability of the training stage. Although we have applied 2-D geometry coherence as one of the criteria during matching individual training samples to their viewpoint-paired synthetic images, it is possible to get wrong matched features at inaccurate positions. Therefore, the semantic part annotations transferred from an individual training image could be far off the ground truth area (e.g., outliers shown by the red circles). With viewpoint consistency, the incorrect annotations are eliminated during aggregation, and our method stably outputs the right position for the targeted semantic parts (e.g., final annotation shown by blue stars) based on the reasonably transferred annotations (e.g., inliers shown by green circles).

\subsubsection{Interpreting Semantic Part Detection}
\label{Experiments:Qualitative:Explainability}

Next, we provide some qualitative results to demonstrate the explainability of our approach. In Figure~\ref{Fig:interpret}, we show examples of how we locate the semantic parts in two image pairs. Each pair includes one testing image and its viewpoint-matched synthetic image.The star represents the location of the semantic parts in each image (learned from training in the synthetic images and got transferred to in the testing images), and the color represents their identity. The transformation is learned using nearby matched features, which are shown by green circles (matched features are linked by red lines). For better visualization, we only display the nearest three features for each semantic part in the figure. This explains what features are used to transfer the annotation from synthetic images to testing images, and helps us understand what is going on during the inference process.

\renewcommand{\figurewidth}{7cm}
\begin{figure}
\centering
\includegraphics[width=\figurewidth]{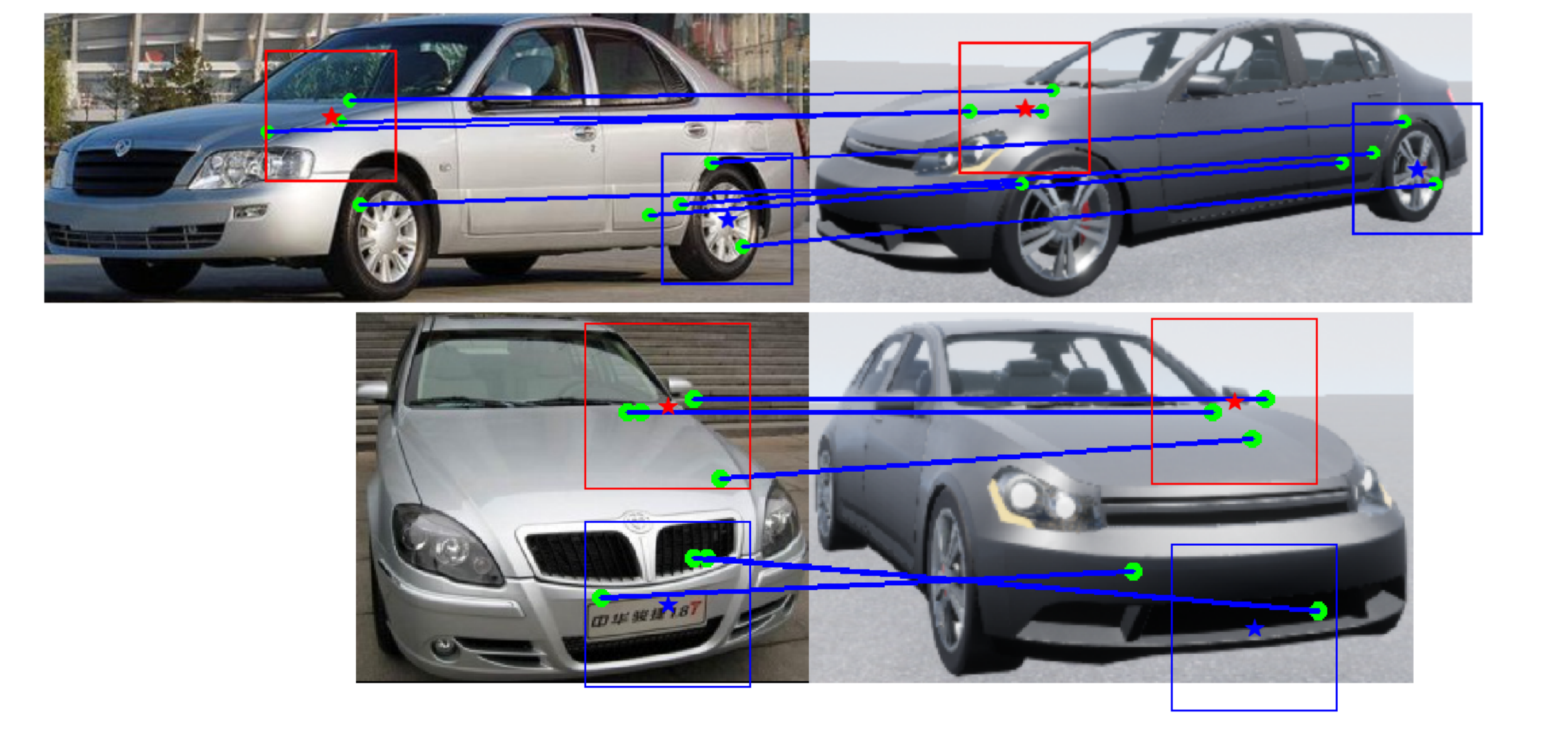}
\caption{Interpret semantic part detection. The matching result is evidence for our approach to detect semantic parts. By visualizing the matching result, we can understand why the approach makes correct or incorrect predictions. \textcolor{black}{Frames represent the detected semantic parts that are transferred from the synthetic image (right) to the testing image (left). Each semantic part is located based on nearby feature matching (shown in blue lines).
}}
\label{Fig:interpret}
\vspace{-0.4cm}
\end{figure}

\section{Conclusions}
\label{Conclusions}

In this paper, we present a novel framework for semantic part detection. The pipeline starts with extracting regional features and applying our robust matching algorithms to find correspondence between images with similar viewpoints. To deal with the problem of limited training data, an additional consistency loss term is added, which measures how semantic part annotations transfer across different viewpoints. By introducing a 3D model as well as its viewpoints as hidden variables, we can optimize the loss function using an iterative algorithm. In the testing stage, we directly apply the same algorithms to match the semantic parts from the 3D model back to each 2D image and achieve high efficiency in the testing stage. \textcolor{black}{Experiments are performed to detect semantic parts of {\em car}, {\em bicycle} and {\em motorbike} in the VSP dataset.} Our approach works especially well with very few ({\em e.g.}, tens of) training images, on which other competitors~\cite{ren2015faster,zhang2018deepvoting} often heavily over-fit the data and generalize badly. 

Our approach provides an alternative solution to object parsing, which has three major advantages: (i) it can be trained on a limited amount of data and generalized to unseen viewpoints; (ii) it can be trained on a subset of viewpoints and then transferred to novel ones; and (iii) it can be assisted by virtual data in both training and testing. However, it still suffers from the difficulty of designing parameters, which is the common weakness of stepwise methods compared to the end-to-end learning methods.

Researchers believe that 3D is the future direction of computer vision. In the intending research, we will try to learn one or more 3D models directly from 2D data, or allow the 3D model to adjust slightly to fit 2D data. More importantly, it is an intriguing yet challenging topic to generalized this idea to non-rigid objects, which will largely extend its area of application.

\section*{Acknowledgement}
This research was supported by ONR grant N00014-18-1-2119 and iARPA DIVA with grant D17PC00342. We thank Huiyu Wang, Yingwei Li and Wei Shen for instructive discussions.
{\small
\bibliographystyle{ieee_fullname}
\bibliography{egbib}
}

\end{document}